\documentclass{article}

\usepackage{arxiv}
\usepackage{tabularx}
\usepackage[toc,title, page]{appendix}

\usepackage[english]{babel}
\usepackage[utf8]{inputenc} 
\usepackage[T1]{fontenc}    
\usepackage{pifont}

\usepackage{csquotes}
\usepackage{amsmath}
\usepackage{graphicx}
  \DeclareGraphicsExtensions{.pdf,.png}
\usepackage{pgfplots}
\pgfplotsset{compat=1.18}
\usepackage{authblk}

\usepackage{biblatex}
\usepackage{amssymb}  
\usepackage{pifont} 
\usepackage{enumitem}
\usepackage[colorlinks=true, allcolors=blue]{hyperref}
\usepackage[most]{tcolorbox}

\usetikzlibrary{positioning, shapes}
\usetikzlibrary{arrows.meta}

\def\BibTeX{{\rm B\kern-.05em{\sc i\kern-.025em b}\kern-.08em
    T\kern-.1667em\lower.7ex\hbox{E}\kern-.125emX}}

\usetikzlibrary{calc}
\usetikzlibrary{fit} 

\makeatletter
\renewcommand\paragraph{\@startsection{paragraph}{4}{\z@}%
  {3.25ex \@plus1ex \@minus.2ex}%
  {-1em}%
  {\normalfont\normalsize\itshape}}
\makeatother

\begin{document}


\title{Triadic Fusion of Cognitive, Functional, and Causal Dimensions for Explainable LLMs: The TAXAL Framework }

\newcommand{\shorttitle}{TAXAL: Triadic Alignment for eXplainability in Agentic LLMs} 

\author{David Herrera-Poyatos$^{1*}$, Carlos Pel\'aez-Gonz\'alez$^1$, Cristina Zuheros$^1$, Virilo Tejedor$^1$, Rosana Montes$^2$, Francisco Herrera$^{1,3}$}

\affil{$^1$Department of Computer Science and Artificial Intelligence, Andalusian Institute of Data Science and Computational Intelligence (DaSCI), University of Granada, Spain. \\ Emails: \texttt{\{divadhp, carlosprog, czuheros\}@ugr.es}, \texttt{virilo@gmail.com}, \texttt{herrera@decsai.ugr.es}
 }
\affil{$^2$Department of Software Engineering, Andalusian Institute of Data Science and Computational Intelligence (DaSCI), University of Granada, Spain. \\ Email:   \texttt{rosana@ugr.es}
}
\affil{$^3$ ADIA Lab, Abu Dhabi,United Arab Emirates\\
$^{*}$Corresponding author.}

\date{\today}

\maketitle
\begin{abstract}
Large Language Models (LLMs) are increasingly being deployed in high-risk domains where opacity, bias, and instability undermine trust and accountability. Traditional explainability methods, focused on surface outputs, do not capture the reasoning pathways, planning logic, and systemic impacts of agentic LLMs. 

We introduce TAXAL (\textit{Triadic Alignment for eXplainability in Agentic LLMs}), a triadic fusion framework that unites three complementary dimensions: \textit{cognitive} (user understanding), \textit{functional} (practical utility), and \textit{causal} (faithful reasoning). TAXAL provides a unified, role-sensitive foundation for designing, evaluating, and deploying explanations in diverse sociotechnical settings. 

Our analysis synthesizes existing methods, ranging from post-hoc attribution and dialogic interfaces to explanation-aware prompting, and situates them within the TAXAL triadic fusion model. We further demonstrate its applicability through case studies in law, education, healthcare, and public services, showing how explanation strategies adapt to institutional constraints and stakeholder roles. 

By combining conceptual clarity with design patterns and deployment pathways, TAXAL advances explainability as a technical and sociotechnical practice, supporting trustworthy and context-sensitive LLM applications in the era of agentic AI.
\end{abstract}


\newpage

\section*{Highlights}
\begin{itemize}
    \item Introduces TAXAL, a triadic fusion framework for explainable LLMs.  
    \item TAXAL integrates three complementary dimensions: \textit{cognitive} (user understanding), \textit{functional} (practical utility), and \textit{causal} (faithful reasoning).
    \item Demonstrates the framework’s applicability through multi-domain case studies (law, education, healthcare, public services).
    \item Proposes design patterns and adoption pathways for practical deployment.  
    \item Positions TAXAL as a scaffold for trustworthy, contestable, and accountable AI.  
\end{itemize}

\section*{Graphical abstract}

\begin{figure*}[ht!]
    \centering
    \includegraphics[width=1\textwidth]{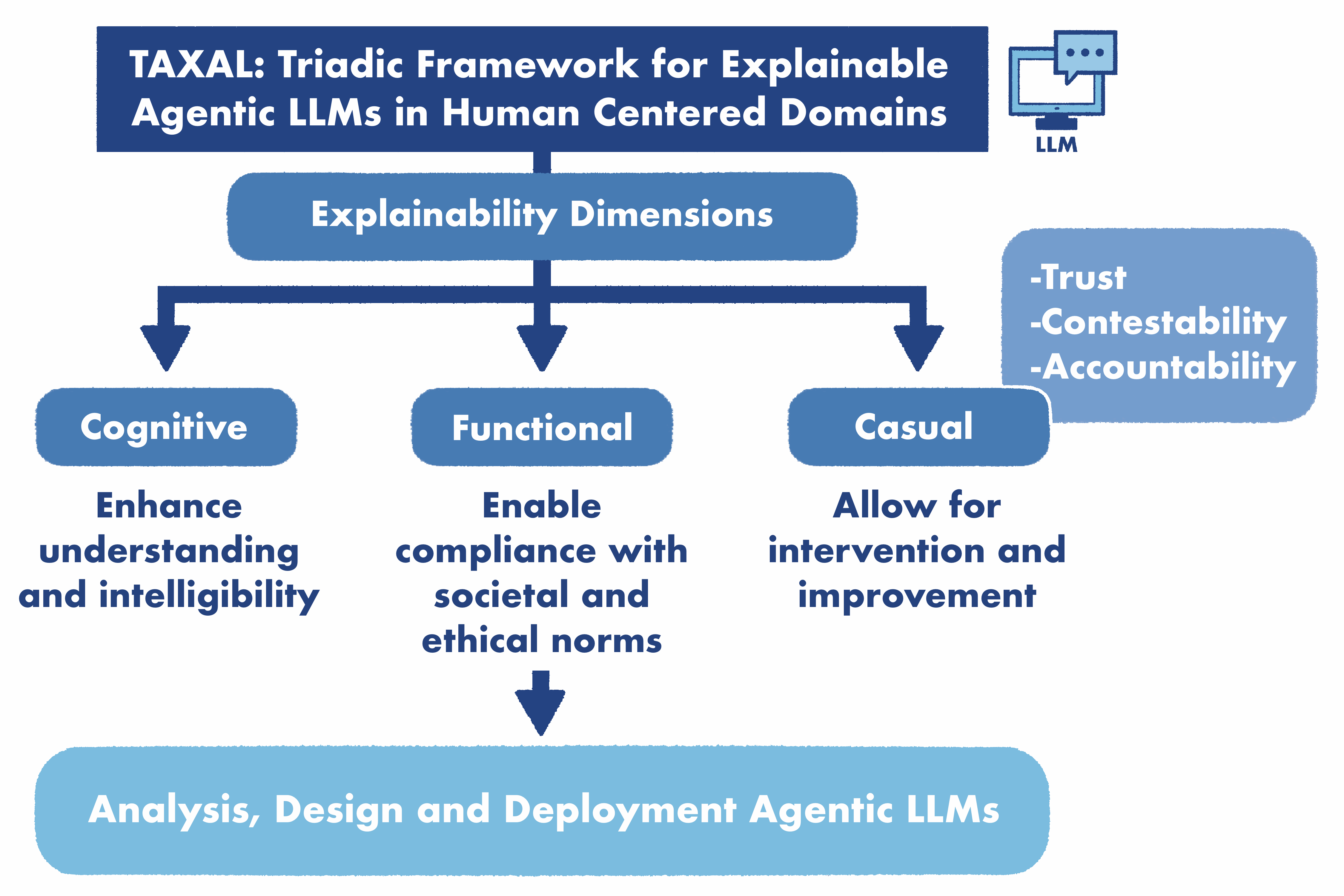}
    \caption{Graphical abstract: TAXAL triadic fusion framework}
    \label{Graphical}
\end{figure*}

\newpage

\section{Introduction}
\label{sec:introduction}

Large Language Models (LLMs) such as GPT-5, GEMINI, Claude, and LLaMA have become foundational tools in artificial intelligence (AI), achieving state-of-the-art performance in summarization, translation, reasoning, and dialogue. However, since LLMs are increasingly integrated in high-risk decision making in domains such as healthcare, law, and education, their lack of transparency raises urgent concerns for safety, accountability, and public trust~\cite{lin2024towards}.

The scale and complexity of these models, covering billions of parameters trained in opaque corpora, make their internal reasoning fundamentally inscrutable. This opacity creates barriers to responsible adoption, as users often lack meaningful ways to understand or challenge outputs. Without stakeholder-sensitive explanations, systems risk overtrust, misinterpretation, or outright rejection~\cite{khalili2024against}.

Explainable AI (XAI) for LLMs has therefore evolved beyond technical introspection~\cite{herrera2025making}. The goal is not only to expose internal mechanisms but also to support human interaction, trust calibration, and decision assurance. As model behavior becomes more emergent and unpredictable~\cite{heyen2024effect}, explanation systems must serve cognitive, functional, and ethical purposes simultaneously~\cite{herrera2025reflections}. Arrieta et al.~\cite{arrieta2020explainable} define explainability as: 

\begin{quote}
\noindent \textbf{Definition.} \textit{Given an audience, an explainable AI is one that produces details or reasons to make its functioning clear or easy to understand.}
\end{quote}

Recent literature underscores a growing consensus that explainability is no longer a peripheral concern but a foundational element of human-centered AI: essential for interpretability, accountability, and regulatory compliance. As emphasized by Cambria et al.~\cite{cambria2024xai}, de Carvalho Souza et al.~\cite{de2025unveiling}, and Mumuni et al. ~\cite{mumuni2025explainable}, explainability must be integrated with the functional capabilities of LLMs, particularly in high-risk and regulated domains. Complementing this perspective, Zhao et al.~\cite{zhao2024explainability} and Zhu et al.~\cite{zhu2024explanation} have identified a critical mismatch between plausible explanations and the faithful behavior of the underlying models, advocating stronger evaluation protocols grounded in human studies and causal reasoning. Zhu et al.~\cite{zhu2024explanation} further propose a unified framework that integrates the formats of free text, structured, and causal explanations. Collectively, these analysis and overviews reflect a broader shift from explanation as static visualization toward explanation as an adaptive, trustworthy dialogue embedded in sociotechnical systems. Shui and Ru~\cite{shui2025bridging} contribute to this evolution by offering a detailed methodological taxonomy of LLM explanation strategies, helping to systematize the growing landscape of tools and approaches.

At the same time, LLMs face systemic risks: jailbreak vulnerabilities~\cite{liu2024jailjudge}, hallucinations and bias~\cite{lin2024towards}, and model uncertainty and variability ~\cite{herrera-poyatos2025overview}. Together, these challenges call for explanation frameworks that are faithful, stakeholder-aware, and resilient to adversarial manipulation and behavioral variability; these capabilities are indispensable as LLMs take on more autonomous and agentic roles.

Crucially, explainability is not monolithic. It varies by stakeholder role: developer, regulator, domain expert, or end user, and is increasingly recognized as essential for legal accountability, fairness, and inclusive governance.  Based on recent trustworthiness concerns, we propose TAXAL (\textit{Triadic Alignment for eXplainability in Agentic LLMs}), a triadic fusion framework that integrates three complementary dimensions: \textit{cognitive} (user understanding), \textit{functional} (practical utility), and \textit{causal} (faithful reasoning). TAXAL connects algorithmic progress in LLMs with role-sensitive explanation strategies, guiding the development of intelligible, trustworthy, and stakeholder-aligned systems.

Our contribution is organized across five axes: (i) \textit{Foundations and evaluation strategies}, consolidating principles, methods, and trade-offs in explainability; (ii) the \textit{TAXAL triadic fusion framework}, structured around cognitive, functional, and causal dimensions; (iii) \textit{Human-centered explanation and interaction}, explored through six use cases in high-risk domains, including a detailed medical scenario; (iv) \textit{Design patterns and adoption pathways}, bridging theory and implementation; and (v) \textit{Research horizons and limitations}, identifying open challenges in validation, contextual generalization, and sociotechnical integration.

Beyond technical introspection, we position TAXAL as the advancement of three essential sociotechnical capabilities for agentic LLMs: \textit{trust}, by aligning explanations with human cognition and institutional expectations; \textit{contestability}, by enabling users and auditors to challenge or refine outputs through counterfactuals and interactive rationales; and \textit{accountability}, by ensuring causal fidelity and audit-ready traces that support compliance and oversight. Together, these capabilities situate triadic fusion not only as a framework for interpretability but as a foundation for a trustworthy, participatory, and ethically governed LLM deployment.

\noindent \textbf{Contribution.} This paper contributes \textit{TAXAL}, a triadic fusion framework that unites the cognitive, functional, and causal dimensions of explainability in LLMs, demonstrated through cross-domain case studies, enriched evaluation strategies, and design patterns that operationalize trustworthy, contestable, and accountable AI in high-risk sociotechnical settings.

The remainder of this paper is structured as follows. Section~\ref{sec:foundations} introduces foundational concepts and evaluation metrics for LLM explainability. Section~\ref{sec:llm_framework} presents the TAXAL framework and maps the explanation strategies to stakeholder roles and scoring matrices. Section~\ref{sec:case} operationalizes the framework through cross-domain applications, including a detailed medical use case. Section~\ref{sec:deployment} outlines the deployment considerations, design patterns, adoption pathways, limitations, and research horizons. Finally, Section~\ref{sec:conclusion} concludes with reflections and future directions. Finally, we include an appendix that provides a broad description of the respective discussions introduced in subsection \ref{sec:axes} and subsection~\ref{sec:strategies} respectively.


\section{Foundations and Evaluation Dimensions}
\label{sec:foundations}

Before we present the TAXAL framework, it is necessary to know and understand the XAI LLMs scenario. Explainability in LLMs encompasses a range of axes and evaluation criteria that influence how models are interpreted and deployed. First, we analyze the definitions of explainability and stakeholder-specific perspectives in Subsection \ref{sec:xai_definitions_stakeholders}.  Subsection \ref{sec:axes} introduces axes of explainability in LLMs proposed by Mumuni et al.\cite{mumuni2025explainable}. Subsection \ref{sec:Holistic} introduces the metrics and evaluation aspects of explainability in LLMs.

\subsection{Explainability Definitions and Stakeholder-Specific Perspectives}
\label{sec:xai_definitions_stakeholders}

The definition and purposes of XAI are not monolithic, but vary between disciplines and stakeholders. According to Arrieta et al. \cite{arrieta2020explainable}, XAI refers broadly to \enquote{the methods and techniques that make the behavior of AI systems understandable to humans}. However, this definition has evolved significantly in the context of LLMs, where scale, emergent behavior, and real-world deployment raise new concerns \cite{herrera2025making}

As emphasized by \cite{herrera2025reflections} and \cite{de2025unveiling}, XAI must go beyond technical introspection and support human and institutional stakeholders in trust calibration, accountability, and societal oversight. This requires tailoring not only the content of the explanations, but also their modality, timing, and presentation format according to the context of use.

Herrera \cite{herrera2025reflections} emphasizes that explainability allows for actionable understanding in the human-AI interaction. The audience, whether an AI developer, a domain expert (e.g., a doctor), or a societal actor (e.g., a patient), critically shapes the explanation goals and formats.

Table~\ref{tab:xai_audiences} summarizes the different audiences for the explainability of LLMs and their corresponding goals, expectations, and challenges.

\begin{table}[ht]
\centering
\caption{XAI Stakeholders, Needs, and Explanation Requirements}
\label{tab:xai_audiences}

\begin{tabularx}{\textwidth}{|X|X|X|}
\hline
\textit{Stakeholder} & \textit{Primary Goal} & \textit{Explanation Requirements} \\
\hline
\textit{Researcher, Designer, Developer} & Debugging and performance alignment & Token-level attributions, internal activations, circuit tracing \\
\hline
\textit{Owner, Regulator or Auditor} & Compliance, fairness, legal auditability & Log trails, counterfactual reasoning, reproducible justifications \\
\hline
\textit{Domain Expert (e.g., Doctor)} & Decision support, safety, traceability & Clear rationale, causal attribution, uncertainty indication \\
\hline
\textit{End User (e.g., Patient)} & Informed consent, fairness, actionability & Simple, culturally accessible explanations; lay terminology \\
\hline
\end{tabularx}
\end{table}

We argue that future XAI frameworks should be aware of the stakeholder by design supporting differentiated explanation pipelines optimized for developer, domain, and social needs. This supports broader adoption in regulated, collaborative, and human-facing LLM deployments.

\subsection{Axes of Explainability in LLMs}
\label{sec:axes}

Following the framework proposed by Mumuni et al.\cite{mumuni2025explainable}, we outline four fundamental axes that structure the landscape of explainability in LLMs. These axes provide a conceptual framework for comparing, selecting and designing XAI strategies in diverse applications.

\begin{itemize}
  \item \textit{Scope:} Global explanations describe the overall behavior of the model (e.g., training dynamics, attention patterns), while local explanations focus on specific outputs or predictions, offering token-level attributions or contextual rationales for a single input.

    \item \textit{Implementation Stage:} Post-hoc methods apply interpretability techniques after model training and prediction (e.g. gradient attribution, saliency mapping). Ante-hoc (inherent) approaches aim to embed interpretability directly into the model design or training objectives, such as through concept bottlenecks or self-rationalizing outputs.
    
    \item \textit{Applicability:} Model-specific methods (e.g., attention visualization, neuron tracing) exploit access to the internal architecture of the LLM. In contrast, model-agnostic approaches (e.g., LIME, SHAP) treat the model as a black box and infer explanations through input-output behavior.

  \item \textit{Opacity and Explainability at Scale:} LLMs operate on opaque training data and contain billions of parameters, making their internal logic inaccessible to symbolic or rule-based inspection. Current explainability techniques approximate model reasoning using proxies like attention weights or saliency maps. However, these are limited in causal insight, especially in highly nonlinear or emergent settings \cite{heyen2024effect}. 
\end{itemize}

These four axes offer a cohesive lens for comparing LLM explainability approaches, emphasizing the trade-offs between scalability, faithfulness, usability, and institutional alignment. These also offer a conceptual framework for classifying and applying different XAI strategies across use cases. The implementation stage, whether explainability is built in (ante-hoc) or applied after the fact (post-hoc), is a key decision. 
   
A detailed description of these four axes can be found in the appendix \ref{app:A}.

\subsection{Evaluation Metrics for Explainability in LLMs}
\label{sec:Holistic}

Evaluating the quality of explanations in LLMs is a complex, multifaceted challenge that goes beyond traditional model accuracy or plausibility checks.

\subsubsection{Faithfulness, Plausibility, Truthfulness and Contrastivity}

Four central criteria guide the evaluation of the explanation \cite{herrera2025making}:

\begin{itemize}
    \item \textbf{Plausibility}: Whether the explanation appears intuitive, reasonable, or convincing to a human user, even if it does not accurately reflect the internal decision-making of the model. High plausibility supports user trust and engagement, but may risk overreliance if not aligned with the model's true logic.
    
    \item \textbf{Faithfulness}: Whether the explanation truthfully represents the actual reasoning process of the model, including the causal or influential factors that led to a specific prediction. Faithful explanations are critical for technical validation, auditing, and compliance in high-risk applications.

 \item \textbf{Truthfulness}: An explanation is truthful if it reflects the source content or external facts. It ensures factual correctness and the absence of hallucinations, which makes it essential in high-risk domains such as medicine or law.

  \item \textbf{Contrastivity}: Tests whether an explanation highlights what distinguishes the chosen output from plausible alternatives, a critical feature for decision support systems. For example, sentiment analysis often distinguishes between positive, neutral, and negative polarities.
  
\end{itemize}

We must note that these evaluation criteria can be mapped directly onto the TAXAL dimensions: faithfulness to the causal axis, plausibility to the cognitive axis, and truthfulness to the functional axis. Contrastivity, in turn, bridges the cognitive and causal axes, since it both clarifies distinctions for users and tests the causal sensitivity of model decisions. In this way, the four criteria align naturally with the triadic fusion structure of TAXAL, offering a multidimensional lens for evaluating explanation quality.

To operationalize these criteria, a range of benchmarks and methods can be used. Each test corresponds to a distinct explanatory goal, whether it is to ensure cognitive plausibility, to verify causal faithfulness, to validate factual truthfulness, or to investigate contrastive reasoning. Representative approaches include:

\begin{itemize}
    \item \textit{Plausibility Tests}: Human evaluations, token masking, or agreement with gold-standard rationales from datasets such as e-SNLI\footnote{https://github.com/OanaMariaCamburu/e-SNLI} or CoS-E\footnote{https://github.com/salesforce/cos-e}.
    \item \textit{Faithfulness Metrics}: Causal tracing, input perturbation, and counterfactual analysis to verify alignment with the model computation.
     \item \textit{Truthfulness Tests}: Benchmarks such as TruthfulQA\footnote{https://github.com/sylinrl/TruthfulQA} or automated judges (e.g., GPT-Judge\footnote{https://chatgpt.com/g/g-T9Vn5BQ7w-judge-gpt}) to assess the factuality of generated explanations.
    \item \textit{Contrastivity Tests}: Benchmarks such as Polyjuice, a general-purpose counterfactual generator\footnote{https://github.com/tongshuangwu/polyjuice} or CELL (a novel adaptive search for contrastive explanations \cite{luss2025cell}.
    
\end{itemize}

These criteria are often in tension. Zhao et al. \cite{zhao2024explainability} contribute a benchmark-driven perspective to LLM explainability by proposing metrics that combine faithfulness, plausibility, and functional utility.  They propose extending standard metrics with dynamic interaction metrics that track changes in user confidence, decision latency, and trust volatility across explanation types. These frameworks bridge model-centric and user-centric perspectives. Notably, they  highlight the importance of evaluating explanations based on task success and human feedback, not just agreement with the ground truth. These suggestions align with our emphasis on moving beyond static plausibility metrics toward outcome-sensitive and interaction-based evaluations.

\subsubsection{Novel Metrics for LLM Explanation Effectiveness}
\label{subsec:mersha_metrics}

Recent literature \cite{zhao2024explainability,zhu2024explanation} emphasizes that the evaluation of explainability in LLMs requires a change from isolated plausibility metrics to user-informed outcomes-driven protocols. This includes triangulating:

\begin{itemize}
    \item \textit{Task-based efficacy:} Whether explanations improve user performance or error detection.
    \item \textit{Behavioral fidelity:} Whether users behave as if they understood the model logic (e.g., via post-explanation prediction).
    \item \textit{Decision impact:} Whether explanations calibrate trust appropriately and influence decisions without overreliance.
\end{itemize}

To go beyond surface-level interpretability scores, Mersha et al.~\cite{mersha2025unified} introduce novel evaluation metrics specifically calibrated for LLM-based XAI methods:

\begin{itemize}
    \item \textit{Human Reasoning Agreement (HRA)}: Measures how closely an explanation matches human rationales, capturing intuitive alignment.
    \item \textit{Robustness}: Assesses the stability of explanations under slight perturbations in the input. Highly volatile explanations indicate poor reliability.
    \item \textit{Consistency}: Ensures that explanations for the same prediction are similar across repeated runs, indicating model determinism and user trustworthiness.
\end{itemize}

These metrics allow for a more comprehensive understanding of explanation quality, particularly in high-risk applications such as healthcare care or legal reasoning.

In summary, explainability in LLMs is not a single-method problem but a multidimensional challenge that must align model behavior with diverse user needs. This is in line with our proposal, the TAXAL triadic fusion framework,  to systematically classify and operationalize explanation strategies in stakeholder contexts.

\section{TAXAL Framework: Triadic Alignment for eXplainability in Agentic LLMs}
\label{sec:llm_framework}

The growing shift from generative to agentic LLMs - systems capable of autonomous reasoning, planning, and interaction - has introduced explainability demands that go far beyond traditional output justification. To address these challenges, we propose the TAXAL framework, which we characterize as a triadic fusion model of explainability. TAXAL integrates three complementary dimensions: cognitive (how explanations align with user understanding), functional (how explanations serve practical decision support and workflow needs), and causal (how explanations trace faithful reasoning pathways). The strength of the framework lies in this fusion: no single dimension is sufficient in isolation, but together they provide a structured, role-sensitive foundation for designing and evaluating explanations that adapt to diverse stakeholders in complex sociotechnical environments.

\subsection*{How to follow TAXAL: A Step-by-Step Guide}

To aid in understanding, we introduce TAXAL not only as a theoretical framework but also as a practical guide for designing, evaluating, and deploying explainability strategies in LLMs. Readers can approach
TAXAL through the following step-by-step flow:

\begin{enumerate}
    \item \textbf{Identify the Stakeholder Role:} Determine whether the explanation is for a developer, regulator, domain expert, or end-user.
    
    \item \textbf{Select the Relevant Dimension:} Map the need for explanation to one of the three axes, \textit{cognitive} (user comprehension), \textit{functional} (practical utility), or \textit{causal}
    (faithful reasoning).
    
    \item \textbf{Choose the Explanation Strategy:} Apply methods aligned with that dimension,
    such as natural-language rationales for cognitive needs, attribution or debugging for functional needs,
    and counterfactuals or tracing for causal requirements.
   
    \item \textbf{Balance Trade-Offs:} Recognize that no single dimension suffices; cognitive clarity,
    functional utility, and causal faithfulness must be fused to meet diverse requirements.
    
    \item \textbf{Iterate in Context:} Test explanations in the actual sociotechnical environment,
    refining them with stakeholder feedback and evaluation metrics.
\end{enumerate}

As an illustrative anchor, consider the medical diagnosis scenario revisited throughout this paper in subsection \ref{sec:medical_use_case}. For patients, TAXAL prioritizes cognitive accessibility through plain-language rationales and functional
decision support for shared care. For doctors, it emphasizes functional diagnostic pathways and
causal counterfactuals for audit and liability. This continuous example highlights how TAXAL
translates abstract principles into a role-sensitive, layered explanation delivery.

\begin{figure}[ht]
\centering
\begin{tikzpicture}[
    node distance=1.6cm and 1.2cm,
    >=Latex,
    font=\small,
    box/.style={draw, rounded corners, align=center, fill=gray!10, minimum width=2.8cm, minimum height=0.9cm},
    mini/.style={draw, rounded corners, align=center, fill=gray!8, minimum width=2.6cm, minimum height=0.8cm},
    thinbox/.style={draw, rounded corners, align=center, fill=gray!6, minimum width=3.0cm, minimum height=0.9cm}
]

\node[box]      (stake) {Identify\\ \textbf{Stakeholder Role}};
\node[box, right=2.6cm of stake] (taxal) {\textbf{TAXAL}\\ Triadic Fusion};

\node[mini, below left=0.9cm and 0.0cm of taxal]  (cog)  {\textbf{Cognitive}\\(user understanding)};
\node[mini, below=0.9cm of taxal]                 (func) {\textbf{Functional}\\(practical utility)};
\node[mini, below right=0.9cm and 0.0cm of taxal] (caus) {\textbf{Causal}\\(faithful reasoning)};


\node[draw, dashed, inner sep=4pt, fit=(cog)(func)(caus)] (triadbox) {};

\node[draw, circle, fill=gray!20, minimum size=5pt, inner sep=0pt, below=1.2cm of func] (merge) {};

\node[thinbox, below=0.4cm of merge] (strategy) {Choose\\ \textbf{Explanation Strategy}};
\node[mini, right=0.8cm of strategy]  (trade) {Balance\\ \textbf{Trade-Offs}};
\node[thinbox, right=0.8cm of trade] (iter) {Iterate in Context\\ \textbf{(Evaluate \& Refine)}};

\draw[->] (stake) -- node[above, align=center]{Scope \& audience} (taxal);
\draw[->] (merge) -- (strategy);
\draw[->] (strategy) -- (trade);
\draw[->] (trade) -- (iter);

\draw[->] (taxal.south west)  to [out=180,in=90] (cog.north);
\draw[->] (taxal.south)       -- (func.north);
\draw[->] (taxal.south east)  to [out=0,in=90] (caus.north);

\draw[->] (cog.south) to [out=270,in=180] (merge.west);
\draw[->] (func.south) -- (merge.north);
\draw[->] (caus.south) to [out=270,in=0] (merge.east);

\node[
  below=0.2cm of triadbox.south,
  fill=white,
  inner sep=2pt,
  rounded corners=1pt,
  text=black
] {%
  \footnotesize Select Relevant \textbf{Dimension}%
};

\node[align=left, anchor=north west] at ([xshift=0.2cm,yshift=-1.0cm]strategy.north west) {e.g., rationales, CoT,\\ attribution, counterfactuals};

\end{tikzpicture}
\caption{\enquote{How to read TAXAL}: from stakeholder role to triadic fusion (cognitive–functional–causal),  then to strategy selection, trade-offs, and contextual iteration.}
\label{fig:taxal_how_to_read}
\end{figure}
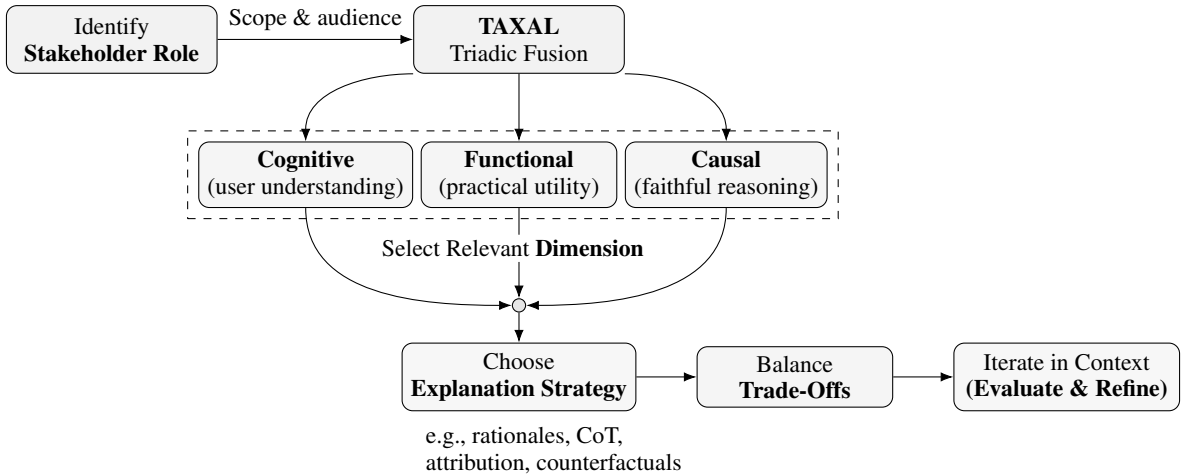

To complement the step-by-step guide, Figure~\ref{fig:taxal_how_to_read} visualizes the TAXAL workflow. 
It begins with identifying the stakeholder role and proceeds through a triadic fusion of cognitive, functional, 
and causal dimensions. These dimensions inform the choice of explanation strategies, which are then balanced
against trade-offs and iteratively refined in context. The diagram highlights how TAXAL structures the path from
abstract principles to practical, role-sensitive explanation design.

In this way, the section is organized to progressively build and validate the proposed TAXAL framework. Subsection~\ref{sec:framework_dimensions} defines the core cognitive, functional and causal dimensions of explainability as applied to LLMs.We show how this framework can be operationalized by mapping it to widely used LLM explanation strategies. Subsection~\ref{sec:strategies} expands this mapping by analyzing the methods that contribute to the alignment of generation, evaluation, and explanation.

\subsection{Cognitive, Functional and Causal Dimensions of Explainability}
\label{sec:framework_dimensions}

TAXAL complements the analysis of the literature on XAI LLMs by introducing a cognitive-functional-causal lens for the use of LLMs in specific applications, agentic LLM. This structure aligns explanation methods not only with their computational basis, but also with user comprehension, functional integration, and causal fidelity across stakeholder contexts. It responds to emerging demands for usability, safety, and human-AI alignment.

Each dimension corresponds to what kind of role an explanation plays.

\begin{itemize}
\item \textit{Cognitive}, how it is understood by humans.

\item \textit{Functional}, how it is used in practice.

\item \textit{Causal}, how faithful it is to the actual behavior of the model.
\end{itemize}

In the following, we develop the three dimensions: 

\paragraph{\textbf{Cognitive Dimension:}} This focuses on how human users perceive, understand, and act on explanations. It includes considerations such as plausibility, comprehensibility, and trust calibration. For example, dialogic and interactive explanations are evaluated not only by their faithfulness to the internal state of the model, but also by how well they align with human mental models and tasks expectations \cite{bertrand2023selective,he2025conversational}. It includes:
    
\begin{itemize}
    \item \textit{Natural language rationales}, which use accessible phrasing to convey decisions or reasoning steps.
    \item \textit{Dialogic explanation interfaces}, allowing users to ask follow-up questions or request clarification in real-time.
    \item \textit{Personalization and cognitive matching}, where explanations adapt to user expertise, background knowledge, or neuro diversity needs \cite{miller2019explanation}.
\end{itemize}

The challenge of making explanations accessible to non-expert users has led to the emergence of audience-adaptive explanation models. Mavrepis et al.~\cite{mavrepis2024xai} propose using LLMs to simplify the outputs of the technical models for diverse audiences. Their system, XAI4All, generates tailored natural language explanations depending on the user's profile, whether a domain expert, policy maker, or layperson.

This approach emphasizes that explainability is not only a property of models but also a communication process. The granularity, terminology, and structure of the explanation must adapt to the cognitive load and goals of the user. Integrating audience-specific prompts or templates in explanation generation can bridge this gap and ensure comprehensibility across user roles in high-risk settings.

Recent empirical findings reinforce this dimension. Wang and Ding~\cite{wang2024rationality} demonstrate that while the presence of explanations can enhance behavioral trust and decision consistency, especially among users with lower task-related capacity—those with higher capacity respond more strongly to the rational quality of explanations. Their study reveals a dual-route processing mechanism: lower-capacity users benefit from the availability of explanations as surface-level cues (a peripheral route), whereas higher-capacity users critically assess the reasoning and evidentiary support behind the explanation (a central route). This distinction highlights the need for cognitively adaptive explanation strategies that provide layered access to information, matching user capacity and informational goals.

\paragraph{\textbf{Functional Dimension:}} This dimension categorizes explanation techniques based on their operational roles across the model lifecycle, from development to deployment and post-deployment monitoring, paying attention to how to use it in practice. Explanations may serve as:

\begin{itemize}
    \item \textit{Diagnostic tools}, which help identify systemic issues such as hallucinations, dataset biases, spurious correlations, or model misalignments. These explanations enable developers to audit system behavior and understand failure modes in a structured, interpretable manner.

    \item \textit{Debugging interfaces}, which surface faulty or incomplete reasoning traces. These tools assist practitioners in tracing the logic behind model outputs, exposing brittle associations, hidden dependencies, or low-reliability inference steps that may not be visible in the raw output.

    \item \textit{User training aids}, designed to scaffold end-user understanding of model predictions, especially in expert domains like medicine, law, or finance. These may include pedagogically structured rationales, interactive \enquote{show your work} steps, or adaptive feedback loops that build trust and improve decision-making literacy.
\end{itemize}

Functional classification binds explainability to practical utility throughout the LLM pipeline, reinforcing its role not just in interpretation, but in iterative system improvement, stakeholder alignment, and risk mitigation \cite{wu2403usable}.

Overall, the functional dimension highlights that explainability is not an afterthought but an integral component of the LLM development pipeline. By serving diagnostic, debugging, fine-tuning, and training roles, explanation methods can improve model robustness, enhance workflow usability, and reduce system-level risks. 

Explanations support both iterative model refinement and stakeholder alignment by surfacing errors early, exposing hidden biases, and enabling adaptive supervision. This perspective encourages the embedding of explainability mechanisms directly into system architectures and development practices, reinforcing their role as core enablers of trustworthy, performance, and context-sensitive LLM applications.

\paragraph{\textbf{Causal Dimension:}} It considers the degree to which an explanation reflects true causal mechanisms within the model. It includes:
\begin{itemize}
    \item \textit{Faithful token attributions}, (e.g., via input perturbation or gradients), which aim to identify which input elements (e.g., tokens or features) truly influenced the model's output. Methods such as input perturbation, integrated gradients, or SHAP attempt to assign importance scores that correlate with output changes, offering localized approximations of causal influence.
    
    \item \textit{Intervention-based tracing} (e.g., causal probing, counterfactuals),   which involves modifying internal or input variables to observe causal effects. Techniques such as causal mediation analysis, neuron activation ablation, or counterfactual input generation help isolate which components or representations drive specific decisions, providing stronger evidence of model behavior under manipulation.

    \item \textit{Post-hoc explanations}, which are constructed after inference and applied to already-trained models. They provide approximations of model reasoning without altering the original architecture. While versatile and widely used (e.g., LIME, SHAP), they may compromise faithfulness since the explanation does not always follow the true rationale of the model’s predictions \cite{mumuni2025explainable}.

    \item \textit{Ante-hoc explanations}, which are embedded during training or model design. These methods aim to endow models with inherent interpretability by learning features and concepts that directly support predictions (e.g., concept bottleneck models). Although often more faithful to causal pathways, they can introduce design constraints that reduce predictive precision \cite{mumuni2025explainable}.
\end{itemize}

    This dimension is particularly important in regulated contexts where auditability, fairness, and compliance are essential.

TAXAL  enables designers and policy makers to align explanation strategies with specific deployment needs. To operationalize these dimensions, we present a unified framework of explanation modes in LLMs (Table~\ref{tab:llm_framework}). The framework organizes explainability along three complementary axes, cognitive, functional, and causal, each associated with representative methods and interpretive goals. Cognitive explanations focus on aligning system outputs with human mental models through natural language rationales or dialogic interaction. Functional explanations emphasize the role of explanations in enhancing model usability, debugging, and robustness within applied workflows. Finally, causal explanations aim to uncover the true decision pathways of the models, providing stronger guarantees of fairness, accountability, and regulatory compliance. Together, these axes offer a structured lens for analyzing and comparing diverse explainability approaches in LLMs.

\begin{table}[ht]
\centering
\caption{A unified framework of explanation modes in LLMs}
\label{tab:llm_framework}
\begin{tabular}{|p{2cm}|p{6cm}|p{6.5cm}|}
\hline
\textbf{Dimension} & \textbf{Representative Methods} & \textbf{Purpose / Interpretation Goal} \\
\hline
\textit{Cognitive} & Natural language rationales, dialogic XAI, progressive disclosure & Align with user mental models; promote trust calibration; adapt to user needs \\
\hline
\textit{Functional} & Hallucination detection, explanation-augmented supervision, LLM-as-evaluator & Improve model robustness, debugging, and usability in workflows \\
\hline
\textit{Causal} & SHAP, LIME, attention rollout, causal tracing, counterfactual reasoning & Trace actual decision pathways; support compliance, fairness, auditability \\
\hline
\end{tabular}
\end{table}

For example, in a medical diagnostic assistant, cognitive clarity is prioritized for patients, while functional decision support and causal reasoning trails are critical for physicians and auditors. A natural-language rationale may suffice for a layperson, but the same system can surface token attributions or counterfactuals for internal auditing. In this way, the framework supports the layered delivery of explanations: cognitively accessible when needed, operationally actionable when demanded, and causally faithful when scrutiny requires it. Rather than treating explainability as a post hoc justification, this model promotes the integration of explanation design throughout the AI development and deployment lifecycle.

 This section illustrates this operationalization by aligning common LLM explainability techniques with each dimension.  Table~\ref{tab:framework-mapping} presents an illustrative mapping of representative explanation techniques in the three dimensions (see a detailed description in \cite{zhao2024explainability} and \cite{shui2025bridging}).

\begin{table}[ht!]
\centering
\caption{Mapping of selected XAI techniques into the LLM explanation Framework}
\label{tab:framework-mapping}
\begin{tabular}{|p{3.8cm}|p{3.5cm}|p{3.5cm}|p{3.8cm}|}
\hline
\textbf{Technique} & \textbf{Cognitive Dimension} & \textbf{Functional Dimension} & \textbf{Causal Dimension} \\
\hline
Chain-of-Thought (CoT) prompting & Transparent reasoning scaffolds & Fine-tuning support, end-user explanation & Weak causal tracing via intermediate steps \\
\hline
Gradient-based attribution (e.g., Grad-CAM) / attention-based attribution (e.g., attention rollout) & Visual salience, faithfulness signals & Safety debugging, hallucination detection & Token-level causal proxies (gradient or flow-based) \\
\hline
Token Attribution (e.g., SHAP\footnote{https://github.com/shap/shap}, LIME\footnote{https://github.com/marcotcr/lime})  & Token-level saliency, plausibility & Debugging, auditing & Approximate causality via perturbation \\

\hline
Counterfactual Explanations & Intuitive reasoning via contrastive “what-if”s & User recourse, fairness auditing, model response exploration & Strong causal alignment via input perturbation and decision boundary testing \\

\hline
Concept Bottlenecks / Interpretable Neural Additive Models (NAMs) & Comprehensibility, symbolic alignment & Ante-hoc interpretability & High causal interpretability \\
\hline
Interactive Dialogic XAI (e.g., TalkToModel) & Dialogic, adaptive, user-aligned & User training, reflective reasoning & Post-hoc natural explanation generation \\
\hline

\end{tabular}
\end{table}

This mapping illustrates that no single technique excels in all dimensions, reinforcing the importance of stakeholder-sensitive explanation strategies that balance cognitive clarity, functional utility, and causal fidelity.

To illustrate how explanation techniques can be evaluated within this framework, we include a representative scoring matrix. Table~\ref{tab:scoring_matrix} shows the alignment strength of the selected techniques using descriptive colors.

\begin{table}[ht]
\centering
\caption{Illustrative mapping of selected XAI techniques to the TAXAL framework dimensions}
\label{tab:scoring_matrix}
\begin{tabular}{|p{4cm}|c|c|c|p{5.5cm}|}
\hline
\textbf{Technique} & \textbf{Cognitive} & \textbf{Functional} & \textbf{Causal} & \textbf{Example Use Case / Remarks} \\
\hline
Chain-of-Thought (CoT) Prompting & \textcolor{red}{High} & \textcolor{red}{High} & \textcolor{orange}{Medium} & Effective for reasoning transparency in QA tasks, but may not trace true causal paths. \\
\hline
Gradient-based attribution (e.g., Grad-CAM) / attention-based attribution (e.g., attention rollout) & \textcolor{orange}{Medium} & \textcolor{red}{High} & \textcolor{orange}{Medium} & Useful for debugging and hallucination analysis, but lacks clear causality. \\
\hline
Token Attribution (e.g., SHAP, LIME) & \textcolor{orange}{Medium} & \textcolor{red}{High} & \textcolor{red}{High} & Strong causal tracing, but cognitive clarity can vary by user expertise. \\

\hline
Counterfactual Explanation  & \textcolor{orange}{Medium} & \textcolor{orange}{Medium} & \textcolor{red}{High} & Causally faithful but may be hard for non-expert users to interpret. \\
\hline
Concept Bottlenecks / Interpretable Neural Additive Models (NAMs) & \textcolor{red}{High} & \textcolor{red}{Medium} & \textcolor{red}{High} & High interpretability but requires model redesign; not always scalable. \\
\hline
Interactive Dialogic XAI (e.g., TalkToModel) & \textcolor{red}{High} & \textcolor{red}{High} & \textcolor{orange}{Medium} & Best for user engagement and adaptation; causal grounding depends on implementation. \\
\hline
\end{tabular}
\end{table}

\begin{itemize}
    \item \textcolor{red}{\textbf{High}}: Strong alignment with the framework dimension.
    \item \textcolor{orange}{\textbf{Medium}}: Moderate or context-sensitive alignment.
    \item \textcolor{green}{\textbf{Low}}: Weak or inconsistent alignment (it does not appear with the current selected XAI techniques). 
\end{itemize}

The utility of TAXAL depends on how well it maps to real-world techniques. Its operationalization bridges the gap between theory and practice and paves the way for LLM explainability tools that are human-centered for different applied high-risk scenarios. We outline potential applications aligned with each dimension.

\begin{itemize}
    \item \textit{Cognitive:} In education, adaptive explanations aligned with student knowledge can enhance learning outcomes and trust in tutoring LLMs.
    \item \textit{Functional:} In biomedical Question/Answer (QA), explanations support the debugging of factual hallucinations while aiding the expansion of the dataset with rational extraction.
    \item \textit{Causal:} In legal or regulatory use cases, intervention-based faithfulness methods (e.g., counterfactual testing) are necessary to ensure compliance and auditability.
\end{itemize}

\subsection{Explanation Strategies in Practice: Generation, Evaluation, and Alignment}
\label{sec:strategies}

 As a complementary study to the previous section, this subsection explores seven explanation strategies currently used to make LLM behavior more explainable and trustworthy. The growing body of research in this area has moved beyond purely theoretical taxonomies, focusing instead on implementable methods that support explanation generation, evaluation, and alignment. 

 A detailed discussion of practical explanation strategies, including natural language rationales, 
conversational interfaces, explanation-enriched prompting, LLMs as evaluators, debugging pipelines, 
and safety prediction mechanisms, is provided in the appendix~\ref{app:strategies}.

On the other hand, to synthesize the variety of strategies explored in this analysis, Table~\ref{tab:llm_framework_mapping} presents a structured mapping between common LLM explainability strategies and the three core dimensions of our framework. This classification highlights how each strategy serves distinct explainability goals. While some approaches (e.g., attribution-based debugging) prioritize internal faithfulness and causal insight, others (e.g., conversational interfaces) emphasize user engagement and adaptive comprehension. 

This mapping offers a clearer view of the trade-offs and design considerations involved in applying XAI tools to LLMs across human-centered applications. This mapping also underscores that current explanation strategies, while diverse, remain fragmented in scope and often optimized for a single dimension.

\begin{table}[ht!]
\centering
\caption{Mapping of LLM explainability strategies to framework dimensions}
\label{tab:llm_framework_mapping}
\begin{tabular}{|p{3.5cm}|p{3.5cm}|p{4cm}|p{4cm}|}
\hline
\textbf{Strategy} & \textbf{Cognitive Dimension} & \textbf{Functional Dimension} & \textbf{Causal Dimension} \\
\hline
\textit{Natural Language Explanations and Self-Rationalization} & High user alignment, accessible to non-experts & Improves user trust and engagement, useful in education and public communication & Limited causal faithfulness, often post-hoc justifications \\
\hline
\textit{Conversational and Interactive Interfaces} & Adaptive to user feedback, supports progressive disclosure & Enhances usability, tailored decision support, scaffolds learning & Post-hoc, low causal depth but flexible for clarification \\
\hline
\textit{Explainable Prompting (e.g., CoT)} & Transparent reasoning scaffolds; aligns with human thought process & Improves zero-shot/few-shot task performance and interpretability & Weak causal tracing via intermediate steps; partially inherent \\
\hline
\textit{LLMs as Explanation Generators/Evaluators} & Simulates expert feedback, provides plausible rationales & Enables annotation, training, and plausibility scoring in low-resource settings & Quality depends on prompt engineering; often lacks ground-truth causality \\
\hline
\textit{Leveraging Explainability for Model Improvement and Debugging}  & Not directly user-facing, but supports developer intuition & Supports model editing, hallucination detection, and factual consistency & High causal grounding through targeted interventions and tracing \\
\hline
\textit{Explainability for Trust, Alignment, and Safety Prediction via Attribution Features} & Indirectly affects user trust via predictive monitoring & Functional utility in safety pipelines, hallucination or bias detection & Weak causal insight; relies on derived features, not model internals \\
\hline
\textit{Model Scale and the Limits of Post-Hoc Explanation} & Higher faithfulness doesn't imply better plausibility & Impacts explanation strategy at deployment scale & Greater traceability but harder to disentangle decision paths \\
\hline
\end{tabular}
\end{table}

A promising direction for future research is to design and deploy explanation tools that explicitly integrate these strategies within the TAXAL triadic fusion framework. Such tools would dynamically align cognitive accessibility, functional utility, and causal faithfulness across stakeholder roles and application contexts.

In practice, this means moving beyond isolated methods, such as attribution maps or conversational interfaces, to composite TAXAL-guided pipelines that adaptively balance interpretability, usability, and auditability. By embedding this integration into real-world workflows, TAXAL can serve not only as a conceptual model, but also as a practical foundation for building explainability-aware LLM systems in high-risk domains.

\section{TAXAL in Action: Cross-Domain Applications and Use Cases}
\label{sec:case}

To analyze the applicability and generalizability of the TAXAL framework in real-world settings, this section presents cross-domain case studies that highlight how cognitive, functional, and causal explanation dimensions interact in practice in subsection \ref{sec:case01}. Each case illustrates a different application context: law, education, public administration, human resources,  mental health screening, and jailbreak detection, where LLMs play high-risk decision-support roles. 
Subsection~\ref{sec:medical_use_case} offers a concrete illustration through a medical diagnosis scenario, highlighting how stakeholder-specific needs shape the explanation content. 

\subsection{Applying TAXAL in Critical Domains}
\label{sec:case01}

\noindent

The following  scenarios are designed not only to map explanation strategies to framework dimensions, but also to surface key trade-offs such as interpretability versus faithfulness or transparency versus cognitive load. In addition, we show how stakeholder-specific needs shape the design and delivery of XAI. These grounded examples provide a concrete foundation for both empirical evaluation and the design of socially aligned, institutionally accountable LLM systems.

Importantly, many of these deployments now feature LLMs with increasing degrees of autonomy and task delegation, hallmarks of agentic behavior. As Schneider outlined~\cite{schneider2025generative}, the shift from generative to agentic AI entails new demands for explainability, including transparency over subgoal decomposition, temporal state tracking, and interactive alignment with human intent. The case studies in this section reflect these evolving demands. Each case illustrates how TAXAL can serve not only as an explanation framework for static LLM outputs but also as a scaffold for layered, adaptive explanation strategies in dynamic, agentic contexts.

A single deployment may require deep, audit-ready causal traces for internal actors (e.g. auditors, domain experts) while also generating simplified, user-facing justifications for nontechnical audiences (e.g. patients, clients, citizens). This dual demand requires adaptive or layered explanation strategies that can accommodate both operational scrutiny and public intelligibility, reinforcing the importance of a context-sensitive explainability design that is stakeholder-aware.


\subsubsection{Use Case 1: Legal Document Review Assistant}

\textbf{Scenario:}

A law firm uses an LLM-based system to assist junior associates in reviewing lengthy commercial contracts. The system flags risky clauses (e.g., indemnification, termination conditions) and suggests edits.\\

\textbf{Framework Mapping:}

Cognitive: Presents plain-language summaries of legal jargon for clarity.

Functional: Serves as a decision-support tool to accelerate clause review and draft iterations.

Causal: Uses counterfactual reasoning to show why a clause was flagged (e.g., \enquote{If this indemnity clause were missing, the contract would pass standard compliance checks}).

\textbf{Trade-offs / Design Tensions:}

Faithfulness vs. Plausibility: Legal justifications must be legally accurate, not just plausible; the risk of hallucination is high.

Comprehensibility vs. Precision: Oversimplifying legal concepts can distort legal meaning; undersimplifying burdens junior users.

\textbf{Internal/External Role Alignment:}

Internal: Legal personnel require high-fidelity causal reasoning and verifiability.

External: Clients may receive simplified risk summaries or compliance explanations on request.

\textit{Commentary:} This case highlights the importance of aligning causal fidelity with legal accountability, while also balancing explanatory clarity for junior staff with readiness to audit compliance. It exemplifies how the stakeholder roles demand different granularity in explanations within the same workflow.

\subsubsection{Use Case 2: Intelligent Tutoring System}

\textbf{Scenario:}

A virtual tutor powered by an LLM helps high school students prepare for math and science exams. It provides hints, evaluates responses, and explains concepts interactively.\\

\textbf{Framework Mapping:}

Cognitive: Tailors feedback based on the student’s knowledge level; adapts tone and language complexity.

Functional: Scaffolds learning by guiding students through problem-solving steps (e.g., progressive disclosure of hints).

Causal: Highlights which parts of the student's answer led to a specific score or correction using token-level saliency.\\

\textbf{Trade-offs / Design Tensions:}

Trust versus Overreliance: Students may accept incorrect but fluent explanations; needs alignment with curricular accuracy.

Adaptive vs. Consistent Feedback: Tailoring explanations dynamically can result in inconsistency across users.\\

\textbf{Internal/External Role Alignment:}

Internal: Teachers may access explanation trails for auditing or grading purposes.

External (Public-facing): Students are non-expert users; priority are clarity, encouragement, and alignment with mental models.

\textit{Commentary:} This example illustrates how cognitive and functional alignment can enhance personalized learning, but must be carefully balanced to avoid overreliance on plausible but pedagogically incorrect rationales. It demonstrates the need for role-sensitive explanation layers that serve both students and educators.


\subsubsection{Use Case 3. Public Service Eligibility Chatbot}

\textbf{Scenario:}

A government agency deploys an LLM-based chatbot to screen citizens for eligibility for housing or unemployment benefits. The chatbot explains decisions based on user-provided inputs.\\

\textbf{Framework Mapping:}

Cognitive: Offers simple justifications for eligibility outcomes (e.g., \enquote{Your income exceeds the current threshold for this program}).

Functional: Supports equitable access to benefits through transparent decision pathways.

Causal: Generates counterfactuals (e.g., \enquote{If your income were below $X$, you would qualify}).\\

\textbf{Trade-offs / Design Tensions:}

Transparency vs. Privacy: Explaining model logic may expose sensitive criteria; must avoid unintended leakage of policy logic.

Comprehensibility vs. Legitimacy: Clear explanations may still provoke distrust if outcomes seem unfair, even if they are correct.\\

\textbf{Internal/External Role Alignment:}

Internal (Auditor/Policy Officer): Internal actors require detailed logs and counterfactual trails for compliance and appeals processes.

External (Citizen): Prioritize ethical clarity, access, and contestability.

\textit{Commentary:} This case emphasizes the ethical imperative of providing transparent yet accessible explanations in citizen-facing systems, while maintaining fairness and regulatory compliance. It showcases the stakeholders are divided between institutional auditability and user-centered clarity in public decision-making.

\subsubsection{Use Case 4: AI-Assisted Hiring Platform}

\textbf{Scenario:}

A company uses an LLM-powered system to help pre-screen job applicants. The system ranks resumes, flags potential mismatches, and generates summary justifications for the review by the recruiter. In some contexts, the same system can offer feedback to rejected applicants on areas of improvement.

\textbf{Framework Mapping:}

Cognitive: Provides recruiter-friendly summaries of candidate strengths and weaknesses, avoiding opaque technical language.

Functional: Streamlines large-scale screening processes, reduces recruiter bias, and supports decision documentation.
  
Causal: Uses counterfactual reasoning to justify rankings (e.g., \enquote{Candidate would have ranked higher with X years of experience in Python}).

\textbf{Trade-offs / Design Tensions:}

Transparency vs. Bias Exposure: Full causal traces may reveal embedded bias or proxy features in training data, raising legal or reputational concerns.

Clarity vs. Legal Risk: Explaining rejection reasons clearly to candidates may help transparency, but introduces liability under employment law.

\textbf{Internal/External Role Alignment:}

Internal: Human resources selection teams and auditors require traceable decision logic and bias mitigation documentation.

External: Applicants benefit from simple, respectful explanations that support transparency and fairness while protecting sensitive model internals.

\textit{Commentary:} This scenario reveals how explainability must navigate legal, reputational and ethical tensions, particularly around bias exposure and applicant fairness. It underscores the dual role of XAI in ensuring both defensible human resource practices and comprehensible and respectful feedback to candidates.

\subsubsection{Use Case 5: Sentiment Analysis in Mental Health Screening}

\textbf{Scenario:}

A mental health platform employs an LLM-based sentiment analysis tool to screen patient entries or therapy chat logs for early signs of depression or anxiety. The system provides risk flags and accompanying explanations to clinicians for classification purposes.

\textbf{Framework Mapping:}

Cognitive: Translates model evaluations into clinician-friendly natural language rationales (e.g., \enquote{Patient expresses loss of interest and hopelessness}).

Functional: Supports early detection and classification by highlighting emotionally significant text regions and changes in sentiment over time.

Causal: Uses contrastive explanations to indicate what made this entry high-risk versus another similar one that was not flagged (e.g., \enquote{Compared to previous entries, this one includes suicidal ideation and reduced self-worth}).

\textbf{Trade-offs / Design Tensions:}

Contrastiveness vs. Sensitivity: Explanations must distinguish between subtle changes in expression (contrastiveness) without oversimplifying complex emotional narratives.

Clarity vs. Clinical Overshadowing: Too detailed causal chains may obscure clinical judgment; too vague may reduce usefulness.

\textbf{Internal/External Role Alignment:}

Internal: Clinical psychologists and triage personnel/staff need robust contrastive and causal explanations to assess risk triggers and monitor the trajectory of sentiment.

External: Patients may receive simplified, non-alarming summaries or feedback (e.g., \enquote{Your recent journal entries show signs of distress - consider checking in with your therapist}).

\textit{Commentary:}  This use case underscores the importance of contrastive and temporally grounded explanations in high-risk applications. In mental health, changes in sentiment are often subtle but critical. Highlighting what specifically differs between flagged and unflagged entry supports informed clinical decisions while managing ethical and emotional sensitivity.


\subsubsection{Use Case 6: Jailbreak Detection and Mitigation in LLM Moderation Systems}

Recent advances in jailbreak detection underscore the growing complexity of safeguarding LLMs from adversarial misuse. Liu et al.~\cite{benchmarkjailjudge} introduce \textsc{JAILJUDGE}, a comprehensive benchmark for evaluating jailbreak resilience in LLMs using a multi-agent, explanation-enhanced framework. Their findings reveal that jailbreak attacks often exploit subtle prompt rephrasings and conversational loopholes, making it essential for safety systems to pair detection mechanisms with interpretable explanations. Motivated by this need, the following case study applies the TAXAL framework to the domain of jailbreak detection, highlighting how cognitive, functional, and causal dimensions interact in the moderation of agentic LLMs under adversarial conditions.

\textbf{Scenario:}

An LLM provider deploys a moderation system to detect and block jailbreak prompts—inputs designed to bypass content filters or elicit harmful responses. When a potential jailbreak is flagged, the system explains its decision to internal reviewers and downstream application teams.

\textbf{Framework Mapping:}

Cognitive: Provides internal reviewers with rationales in natural language explaining what prompt structures, tone changes, or adversarial cues triggered the classification (e.g. \enquote{Prompt contains covert instruction patterns common in known jailbreaks}).
    
Functional: Supports moderation workflows by flagging high-risk prompts and enabling appeal or refinement. Allows system developers to update filters based on explanation feedback.
    
Causal: Traces the decision to specific input perturbations or token combinations that cause the model to deviate from its alignment objectives. May leverage causal mediation or feature ablation to justify high-confidence detections.

\textbf{Trade-offs / Design Tensions:}

Transparency vs. Exploitability: Too much causal detail in explanations may guide attackers in crafting more sophisticated jailbreaks. Explanations must balance clarity with obfuscation.

Auditability vs. Operational Latency: Deep causal traces can be valuable for audits, but may slow down moderation systems in real-time applications.

\textbf{Internal/External Role Alignment:}

Internal: Security analysts and developers require traceable causal paths and model behavior patterns to improve filter robustness and perform root-cause analysis.

External: In limited public or partner-facing contexts, simplified explanations can inform API users of rejected prompts without revealing sensitive detection logic (e.g., \enquote{Prompt violated safety guidelines on prohibited behavior requests}).

\textbf{Commentary:} This use case highlights the need for role-sensitive and adversarially robust explanation strategies in agentic AI safety systems. The cognitive-functional-causal axes help moderation teams navigate the trade-off between intelligibility and system hardening, supporting both actionable defense and post hoc analysis.\\

\noindent
Table~\ref{tab:triadic-case-studies} summarizes the case studies discussed above by systematically mapping each domain to the cognitive, functional, and causal dimensions of the TAXAL framework. It highlights how explanation strategies are tailored to the roles and expectations of different stakeholders, ranging from legal professionals to students, public service, and human resources users, while also illustrating the types of causal insight required in each scenario. The inclusion of internal/external roles further clarifies the dual focus on internal model accountability and external intelligibility. This comparative view supports the operational relevance of the framework across heterogeneous sociotechnical contexts.

\begin{table}[ht]
\centering
\caption{Cross-domain application of the TAXAL framework}
\label{tab:triadic-case-studies}
\resizebox{\columnwidth}{!}{%
\begin{tabular}{|p{2.5cm}|p{1.7cm}|p{2.5cm}|p{2.5cm}|p{2.3cm}|p{2.5cm}|}
\hline
\textbf{Case Domain} & \textbf{User Role} & \textbf{Cognitive Focus} & \textbf{Functional Use} & \textbf{Causal Insight Level} & \textbf{ Internal/External Role} \\
\hline
Legal Document Review & Associate, Client & Legal term simplification & Risk flagging, clause rewriting & Counterfactuals & Internal (developer),  External(client) \\
\hline
Education Tutoring System & Student, Teacher & Progressive scaffolding, clarity & Interactive learning, hinting & Saliency + examples & Internal (teacher), External (student)  \\
\hline
Public Service Eligibility & Citizen, Auditor & Eligibility logic, fairness cues & Benefit screening, appeals & Input sensitivity & Internal(auditor), External (citizen),  \\
\hline
Human Resources & Applicants, Selection Team & Summary feedback on strengths/weaknesses & Resume screening, bias reduction, auditability & Counterfactuals (e.g., feature-based ranking changes) & Internal (HR team), External (applicants) \\
\hline
Mental Health Sentiment Alanlysis & Patient, Clinician & Nature language rationales for risk & Sentiment shift monitoring, early risk flagging & Contrastive explanations (temporal and semantic) & Internal (clinician), External (patient) \\
\hline
Jailbreak Detection & Security Team, API Users & Prompt structure rationale, adversarial pattern detection & Moderation workflows, appeal handling & Token-level perturbation tracing, causal probes & Internal (moderators), External (API clients) \\
\hline
\end{tabular}
}
\end{table}

These case studies operationalize the TAXAL framework by translating abstract dimensions, cognitive, functional, and causal, into concrete explanation strategies that respond to real-world sociotechnical demands. In doing so, they demonstrate how explainability must be adapted not only to the capabilities of LLMs but also to the epistemic needs, knowledge levels, and responsibilities of distinct stakeholder groups.

For example, the legal domain foregrounds the importance of causal fidelity and auditability for compliance, while education emphasizes cognitive clarity and pedagogical alignment. Public administration, on the contrary, illustrates the tension between fairness, transparency, and accessibility in services aimed at citizens. By situating explanation design within the roles of internal, expert-facing, and external, public-facing, these examples show how the same LLM agent system may need to support dual explanation modes: technical depth for auditors and plain language for end users. 

This dual orientation is essential for building not only trustworthy but also socially legitimate AI systems. Ultimately, these use cases validate the framework's capacity to scaffold the design of explainable LLM pipelines that are domain sensitive, role-aware, and aligned with institutional accountability and human-centered deployment goals.

\subsection{Triadic Fusion in Medical Diagnosis (Experts vs. End-Users)}
\label{sec:medical_use_case}

Recent advances in agent-driven LLM AI have underscored the need for explanation strategies that are tailored to the needs of diverse stakeholders, particularly in high-risk domains such as healthcare. As Schneider~\cite{schneider2025generative} argues, agentic systems must move beyond static, one-size-fits-all explanations to deliver context-aware, role-sensitive rationales that support both decision assurance and institutional accountability.

This use case illustrates how the triadic fusion of cognitive, functional, and causal explanations can be adapted to stakeholder roles in medical diagnosis, employing differentiated strategies for internal stakeholders such as physicians and external stakeholders such as patients.

To concretely demonstrate the application of this framework in a human-centered high-risk domain, we consider a clinical decision support system (CDSS) powered by an LLM for diagnosing cardiac conditions from electronic health records (EHR) and physician notes.

In this scenario, the LLM processes a patient's medical history, lab results, physician documentation, and narrative input, returning a preliminary diagnosis with a confidence score. Crucially, the form and depth of the explanation must vary based on the intended audience, whether a medical professional (e.g., cardiologist) or a nonexpert patient, highlighting the need for stakeholder-specific explainability.

\paragraph{Doctor-Centered Explanation.}
For clinicians, the explanation should support rigorous medical reasoning and model critique. This includes:
\begin{itemize}
    \item \textit{Cognitive Dimension:} Highlighting salient clinical markers (e.g. \enquote{elevated troponin}, \enquote{ST-segment elevation}) that influenced the model's output.
    \item \textit{Functional Dimension:} Presenting diagnostic reasoning paths similar to Chain-of-Thought prompts (e.g. \enquote{Given the ECG findings and patient history, myocardial infarction is most likely}).
    \item \textit{Causal Dimension:} Enabling auditability through counterfactuals (e.g., \enquote{If the ECG had not shown ST-elevation, the diagnosis would shift to unstable angina}).
\end{itemize}

\paragraph{Patient-Centered Explanation.}
For patients, the priority shifts to comprehensibility and reassurance:

\begin{itemize}
    \item \textit{Cognitive Dimension:} Providing plain-language rationales (e.g., \enquote{You have moderate asthma and the AI recommends the blue inhaler.}).
    \item \textit{Functional Dimension:} Supporting shared decision making by describing care options and encouraging questions (e.g., \enquote{This inhaler helps reduce inflammation and ease breathing during attacks}). 
    \item \textit{Causal Dimension:} Ensuring transparency and fairness without overwhelming technical detail (e.g., \enquote{If your asthma were more severe, a steroid-based option could be chosen instead.}).
\end{itemize}

As shown in Table~\ref{tab:medical_framework}, different stakeholders emphasize different dimensions of the TAXAL framework depending on their roles and goals. Internal actors such as physicians typically require detailed multilayered explanations that align with clinical reasoning (cognitive), support verification and oversight (functional), and enable causal interrogation through techniques like counterfactuals and tracing. In contrast, external stakeholders, such as patients, benefit most from cognitively accessible rationales that build trust and support informed consent, even if the causal depth is abstracted or simplified. When framed in terms of fairness or legitimacy, even lightweight counterfactuals and transparent decision rationales can meaningfully enhance user understanding and reassurance in sensitive domains like healthcare.

\begin{table}[ht]
\centering
\caption{Framework dimensions emphasized by stakeholder}
\label{tab:medical_framework}
\begin{tabular}{|p{1.5cm}|p{4.2cm}|p{4cm}|p{4.4cm}|}

\hline
\textbf{Audience} & \textbf{Cognitive Focus} & \textbf{Functional Need} & \textbf{Causal Depth} \\
\hline
Doctor & Clinical reasoning alignment & Model critique, verification & Causal tracing, counterfactuals \\
\hline
Patient & Understandable rationale, reassurance & Informed consent, trust-building & Fairness, legitimacy \\
\hline
\end{tabular}
\end{table}

\begin{figure}[ht]
\centering
\begin{tikzpicture}[node distance=1.7cm, >=latex, thick]

\node[draw, rounded corners, align=center, fill=blue!10, minimum width=2.8cm, minimum height=1.2cm] (patient) {Patient \\ (External Stakeholder)};

\node[draw, rounded corners, align=center, fill=green!10, minimum width=2.8cm, minimum height=1.2cm, right=6cm of patient] (doctor) {Doctor \\ (Internal Stakeholder)};

\node[draw, rounded corners, align=center, fill=orange!15, minimum width=12cm, minimum height=1.4cm, below=3cm of $(patient)!0.5!(doctor)$] (cdss) {Clinical Decision Support System (LLM-based)};

\draw[->] (patient.south) -- node[left, xshift=0.1cm, align=center] {Symptoms,\\ patient history} (cdss.north -| patient);
\draw[->] (doctor.south) -- node[right, xshift=-0.1cm, align=center] {EHRs, labs,\\ clinical notes} (cdss.north -| doctor);

\draw[->] (cdss.north -| patient) -- (patient.south) node[midway, left, align=left, xshift=4.4cm] {
    \textbf{Patient-facing Explanation:}\\
    \scriptsize Cognitive: Plain-language rationale\\
    \scriptsize Functional: Shared decision support\\
    \scriptsize Causal: Fairness-level counterfactuals};

\draw[->] (cdss.north -| doctor) -- (doctor.south) node[midway, right, align=left, xshift=-4.4cm] {
    \textbf{Doctor-facing Explanation:}\\
    \scriptsize Cognitive: Salient clinical markers\\
    \scriptsize Functional: Diagnostic reasoning path\\
    \scriptsize Causal: Counterfactual tracing};

\end{tikzpicture}
\caption{Application of the TAXAL triadic fusion framework in a clinical decision support system. The same LLM-driven CDSS produces layered explanations adapted to stakeholder roles: simplified narratives for patients, and detailed causal reasoning for physicians.}
\label{fig:taxal_medical_use_case}
\end{figure}
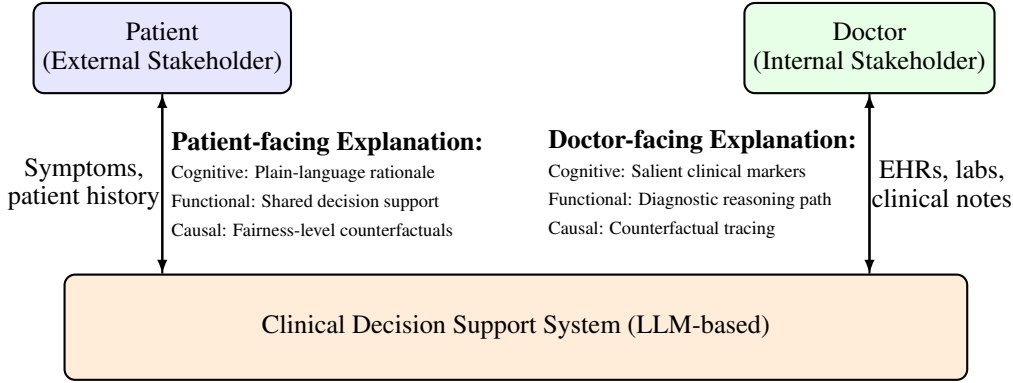

To further illustrate how TAXAL can be operationalized in a real clinical workflow, Figure~\ref{fig:taxal_medical_use_case} shows a role-sensitive explanation pipeline within a CDSS. The same LLM processes patient symptoms, EHRs, and clinical notes, but generates explanations tailored to stakeholder roles. Patients receive simplified, cognitively accessible rationales that support trust and informed consent, while doctors receive layered reasoning paths and counterfactual traces that enable verification, critique, and liability documentation. This diagram emphasizes the central contribution of TAXAL: aligning the depth and modality of the explanation with the needs of heterogeneous stakeholders in high-risk domains.

This use case illustrates that different dimensions may sometimes conflict: for example, counterfactuals may aid doctors but confuse patients. Explanation strategies should therefore be \emph{dynamically tailored}, balancing faithfulness with cognitive accessibility.

A more concrete scenario highlights how TAXAL works in practice. Consider an emergency room case where a patient presents with acute chest pain. The CDSS processes the EHR data, laboratory tests, and the ECG, suggesting a high probability of myocardial infarction. For the cardiologist, the system generates a layered explanation: it highlights elevated troponin levels and ST-segment abnormalities (cognitive), reconstructs its reasoning path through sequential findings (functional), and provides a counterfactual analysis showing how the absence of ST elevation would lower the probability of infarction (causal). In contrast, for the patient, the system produces a simplified narrative: \enquote{Your test shows a pattern often linked to a heart attack, and immediate treatment is recommended}. Here, causal information is communicated at a level of fairness and legitimacy, avoiding technical overload while still supporting informed consent. 

This scenario illustrates how the same CDSS can deliver explanations aligned with stakeholder roles without duplicating effort. The doctor’s interface integrates interactive modules to drill down to token-level contributions or run \enquote{what if} simulations, enabling clinical verification and liability documentation. Meanwhile, the patient interface emphasizes trust calibration, combining accessible rationales with reassurance, and explicitly invites shared decision-making questions. These dual explanation layers showcase TAXAL’s capacity to reconcile technical auditability with cognitive accessibility in a single deployment pipeline.

\paragraph{Toward Empirical Validation of the Framework.}
The above scenario also raises important questions about how explanation methods align with user needs in practice. To rigorously evaluate and compare such methods, we propose an TAXAL-based validation framework.

\begin{tcolorbox}[colback=gray!5,colframe=gray!80,title=Guidance: Empirical Validation Strategy]
\begin{itemize}
    \item \textbf{Apply}: Deploy the framework in real-world tasks (e.g., medical QA, legal clause review, or model debugging).
    \item \textbf{Measure}: Use stakeholder-centered metrics such as trust calibration, comprehension accuracy, and counterfactual plausibility.
    \item \textbf{Score}: Develop matrices that map the explanation techniques to the cognitive, functional, and causal dimensions.
\end{itemize}
\end{tcolorbox}

\noindent \textbf{Remark:} We must note that this is a simulated scenario without real doctors and patients. Future work is to design explainable LLM pipelines (specific instances of TAXAL) in a real scenario with stakeholders that are domain sensitive, role aware, and aligned with institutional accountability and human-centered deployment goals. 

In the following section, we propose this future development as a guide for the design and deployment of explainable LLM pipelines. 


\section{Deploying Triadic Fusion: Considerations, Patterns, Adoption Pathways, Limitations and Research Directions}
\label{sec:deployment}

Although TAXAL offers a conceptual and evaluative structure for explainability, its real-world impact depends on its integration into the design and operational lifecycle of LLM-powered systems with an explainable LLM pipeline, particularly those that exhibit agentic behaviors such as autonomous reasoning, tool use, and stakeholder interaction.

This section outlines considerations from the use cases analysis. Then we analyze design patterns and adoption strategies that can help practitioners implement stakeholder-aligned explainability in practical high-risk settings. Finally, limitations and future research are discussed. 

\subsection{TAXAL Considerations}

\noindent
The studies presented above yield several overarching insights relevant to the design, deployment, and oversight of explainable LLM systems, especially as they evolve toward increasingly agentic behavior. These reflections highlight how stakeholder-sensitive explainability strategies must be context-aware and governance-aligned. 

\begin{itemize}
    \item \textbf{Layered stakeholders explanation strategies are essential}: Systems must often support both internal and external roles simultaneously, requiring multilevel explanation formats that range from technical audit trails to accessible rationales.
    
    \item \textbf{Explanation design must reflect the domain-specific constraints}: The legal, health, educational, and public service domains each must impose unique requirements on the four concepts—\textit{faithfulness}, \textit{Truthfulness}, \textit{plausibility}, and \textit{contrastivity}, necessitating different trade-offs and presentation modalities.
    
    \item \textbf{Causal reasoning is critical for compliance and contestability}: Across domains, counterfactuals and traceable attribution emerged as necessary tools to satisfy both legal accountability and ethical transparency.
    
    \item \textbf{Public-facing systems must prioritize fairness and intelligibility}: In domains such as education and citizen services, the clarity, tone, and perceived legitimacy of explanations play a decisive role in building user trust and supporting informed consent.
\end{itemize}

\subsection{Explanation Design Patterns}

To support the cognitive, functional, and causal dimensions across internal/external user roles, we propose a set of explanation design patterns as guidelines:

\begin{itemize}
    \item \textbf{Layered Explanation Interfaces:} Systems should offer tiered explanation layers, e.g., surface-level rationales for end-users (external) and in-depth causal traces or model justifications for auditors (internal).
    
    \item \textbf{Explanation Role Routing:} Role-aware interfaces can dynamically adapt explanation granularity based on user profiles, e.g., regulators see compliance justification logs, while consumers receive accessible summaries.

    \item \textbf{Counterfactual Feedback Modules:} Modular components can simulate \enquote{what-if} scenarios, offering users interpretable ways to explore decision boundaries (e.g., resume scoring, eligibility thresholds).

    \item \textbf{Explanation-aware Prompt Templates:} In LLM workflows, prompt engineering should explicitly embed the triadic dimensions (e.g., cognitive summarization, functional reasoning steps, causal sensitivity triggers).
\end{itemize}

\subsection{Technical Adoption Pathways}

To deploy TAXAL in production systems, we propose a set of guide requirements:

\begin{itemize}
    \item \textbf{Explainability-Aware Documentation:} System cards, model cards, and audit trails should explicitly refer to which dimensions (cognitive, functional, causal) are addressed and for whom.

    \item \textbf{Procurement and Compliance Integration:} Government and enterprise procurement pipelines should include triadic-based explainability criteria, aligned with internal / external XAI roles to meet transparency regulations.

    \item \textbf{Participatory Feedback Loops:} Explanation effectiveness should be evaluated iteratively with stakeholders, including non-expert users, to ensure cognitive accessibility and trust calibration.
\end{itemize}


\subsection{Limitations and Research Horizons for TAXAL in Agentic LLMs}
\label{sec:Limitations}

As LLMs evolve toward increasingly agentic capabilities—autonomous planning, tool use, and adaptive interaction—the demands on explanation frameworks such as TAXAL intensify. Although the cognitive-functional-causal triad provides a robust scaffold for aligning explanation strategies with human understanding, task utility, and causal fidelity, emerging use cases expose conceptual and practical needs. 

This section critically reflects on these needs, discussed as limitations and future research, ranging from empirical validation and stakeholder generalization to sociotechnical integration and agentic system explainability. 

We identify key frontiers for research and design innovation, advocating for a shift from static taxonomies to adaptive, dynamic, and governance-aware XAI ecosystems that can sustain trust and traceability in increasingly complex LLM deployments. To avoid a long description of them, we propose it as a short summary as a table.  Table~\ref{tab:framework_limitations} discusses these limitations in both conceptual and practical terms to explain the research horizons to apply framework-based XAI to LLM, while also highlighting clear avenues for empirical refinement, task-specific adaptation and sociotechnical integration.

\begin{table}[ht]
\centering
\caption{Summary of limitations and research horizons in framework-driven XAI}
\label{tab:framework_limitations}
\begin{tabular}{|p{4cm}|p{6cm}|p{5cm}|}
\hline
\textbf{Limitation} & \textbf{Description} & \textbf{Future Direction} \\
\hline
Contextual Generalization & Framework lacks validation across real-world domains and tasks. & Conduct domain-specific studies to test explanation modes under varying user and task constraints. \\
\hline
Lack of Formal Mapping Criteria & Current mapping relies on qualitative judgment. & Develop standardized scoring rubrics for cognitive, functional, and causal alignment. \\
\hline
Empirical Evaluation Gaps & No quantitative assessment across metrics or datasets. & Apply the framework using benchmarks (e.g., ERASER, TruthfulQA) to validate claims. \\

\hline
Evolving Explanation Modalities & New paradigms (e.g., multimodal or agentic XAI) stretch existing dimensions. & Extend framework to support hybrid or emergent explanation strategies. \\
\hline
Need for Sociotechnical Integration & Current focus is technical, neglecting governance and ethics. & Incorporate socio-political dimensions, regulatory contexts, and power asymmetries. \\
\hline
Faithfulness vs. Plausibility Tradeoff & Tradeoff between technical accuracy and user belief remains unresolved. & Develop methods balancing causal correctness and communicative clarity. \\
\hline
Task-Specific Limits & Some methods perform poorly on semantically complex tasks. & Design task-sensitive evaluation protocols tailored to task characteristics. \\
\hline
Limited Support for Agentic Dynamics & Framework does not account for goal formation, subtask decomposition, or multi-agent coordination in agentic systems. & Extend framework to include temporal, systemic, and multi-agent explanation structures. \\

\hline
Agentic System Demands & Framework lacks mechanisms to trace dynamic intent, delegation, and inter-agent interaction. & Extend dimensions with temporal traces, goal provenance graphs, and action-based causal justifications. \\

\hline
\end{tabular}
\end{table}

 To bridge the conceptual underpinnings of TAXAL with its practical implementation, future research must translate its principles into scalable tools and evaluative infrastructure. We will work as future researchers to treat this framework not as a prescriptive one but as a generative tool for interdisciplinary dialogue and iterative improvement(see studies in \cite{bertrand2023selective,he2025conversational}). 

\section{Conclusion}
\label{sec:conclusion}

This paper has introduced TAXAL, a triadic fusion framework that integrates cognitive clarity, functional utility, and causal faithfulness into a role-sensitive foundation for explainability in agentic LLMs. Our central claim is that no single dimension is sufficient in isolation; only through their integration can explanation strategies meet the diverse needs of stakeholders in high-risk sociotechnical domains.

We substantiated this claim on three levels: (i) by synthesizing and classifying existing explainability methods within the cognitive–functional–causal structure, (ii) by operationalizing the framework through six cross-domain case studies and a detailed medical diagnosis scenario, and (iii) by translating the deployment analysis into design patterns and adoption pathways that bridge conceptual taxonomies with practical deployment. 

Looking ahead, TAXAL serves as a scaffold methodology for adaptive, governance-aware XAI LLM ecosystems. Embedding triadic fusion into the LLM design, evaluation, and oversight lifecycle allows explanation strategies that go beyond surface plausibility to support three essential capabilities of responsible AI: \textit{trust}, through cognitively accessible rationales; \textit{contestability}, through counterfactuals and role-sensitive justifications that empower users to challenge outputs; and \textit{accountability}, through causal traces and audit-ready explanations that ensure compliance and institutional oversight. 

In this way, TAXAL reframes explainability as a shared language of trust between humans and machines and as a practical scaffold for transparent, participatory, and ethically aligned AI systems. The following highlights distill its contributions and future directions.

\begin{tcolorbox}[colback=gray!5,colframe=gray!80,title=Key Takeaways]
\begin{itemize}
    \item \textbf{Triadic Fusion:} Introduced TAXAL, a framework that integrates cognitive, functional, and causal dimensions for role-sensitive explainability in LLMs.  
    \item \textbf{Cross-Domain Relevance:} Demonstrated applicability through six case studies and a detailed medical use case, showing how explanations adapt to diverse stakeholders.  
    \item \textbf{From Theory to Practice:} Proposed design patterns and adoption pathways that operationalize TAXAL deployment in high-risk sociotechnical and real-world settings.  
    \item \textbf{Future Direction:} Positioned TAXAL as a scaffold methodology to build adaptive, governance-aware XAI ecosystems, enabling trustworthy, contestable, and accountable explanations in agentic AI.  
\end{itemize}
\end{tcolorbox}

\begin{appendices}
\section*{Description Axes of Explainability  and Explanation Strategies}\label{app:desc}
We include two appendices providing a wide description of the respective discussions introduced in subsection \ref{sec:axes}   and subsection~\ref{sec:strategies} respectively. 

\section{Additional Description Axes of Explainability  in LLMs}
\label{app:A}

This appendix provides extended discussion of four core axes of explainability in LLMs discussed in subsection \ref{sec:axes}  proposed by Mumuni et al.\cite{mumuni2025explainable}. Each subappendix (1-4) expands a deep analysis on theoretical and practical perspectives of each one.

\subsection{Local vs Global Explanation Techniques}
\label{app:scope}

Explanation methods can be categorized by their analytic scope \cite{luo2024understanding}:

\begin{itemize}
    \item \textit{Local Explanations}: Focus on individual predictions. Examples include token saliency, attention rollout, and local perturbation analysis (e.g., LIME).
    \item \textit{Global Explanations}: Provide insights into general model behavior, such as probing classifiers, representation clustering, or circuit tracing. See a recent study for circuit tracing by Rai et al.~\cite{rai2025practical}
\end{itemize}

These approaches serve different user needs: debugging and transparency for local users versus auditability and fairness for global stakeholders.

\subsection{Implementation: Post-hoc vs Ante-hoc Methods}
\label{app:implementation}

Explainability methods for LLMs generally fall into two categories: \textit{inherent} (ante-hoc) and \textit{post-hoc}. Although most current approaches are post-hoc—applied after model training and inference—there is growing interest in developing models with built-in, or \textit{inherent}, interpretability as a core design objective \cite{mumuni2025explainable}.

\paragraph{Post-Hoc Explainability.}  
Post-hoc methods seek to extract explanations from models already trained without altering their internal mechanisms. These include:

\begin{itemize}
    \item \textit{Token-level attribution} (e.g., LIME, SHAP): These techniques estimate the contribution of individual input tokens to the output. Although popular for their model-agnostic nature, they often assume linear separability and suffer from context sensitivity, particularly in autoregressive architectures such as LLMs \cite{heyen2024effect}.

    \item \textit{Attention-based visualizations}: Heat-maps of a Transformer’s attention weights that highlight which input tokens the model \enquote{looks at}; however, these weights capture correlation rather than causal influence, so high attention does not necessarily mean a token drives the prediction—hence the ongoing debate over their faithfulness.

    \item \textit{Explanation generation:} Some approaches prompt LLMs to produce natural language justifications. However, these may reflect plausible narratives rather than actual reasoning paths, risking overtrust due to high fluency.
\end{itemize}

Post-hoc techniques are especially useful for auditing and debugging black-box models, but they often lack faithfulness and are limited in their ability to explain emergent behaviors in large-scale systems.

\paragraph{Inherent (Ante-Hoc) Explainability.}  
In contrast, \textit{inherent} or \textit{ante-hoc} explainability involves designing models to be transparent by construction. Examples include:

\begin{itemize}
    \item \textit{Concept Bottleneck Models (CBMs):} These models first predict human-understandable concepts and then use those to generate the final output, ensuring interpretability throughout the pipeline.
    \item \textit{Neural Additive Models (NAMs):} These extend generalized additive models to deep learning settings, retaining interpretability while leveraging non-linearity.
    \item \textit{Self-rationalizing architectures:} These models are trained to jointly predict outputs and generate faithful rationales as supervision signals, improving the alignment between internal decision-making and explanation.
\end{itemize}

Although inherently explainability models offer greater transparency and potentially greater faithfulness, they often come at the cost of scalability and predictive power. As noted by Mumuni et al. \cite{mumuni2025explainable}, the trade-off between model accuracy and interpretability remains a central tension in the explainable LLM design.

\paragraph{Hybrid Directions.}  
Emerging research explores hybrid approaches that embed explanation constraints during fine-tuning (e.g., rationale supervision) or incorporate transparent modules into otherwise opaque LLMs. These methods aim to combine the scalability of black-box architectures with interpretable scaffolding, offering a promising path for explainability-by-design in future LLM development \cite{zhao2024explainability, cambria2024xai}.

In summary, the distinction between inherent and post-hoc XAI reflects not only methodological choices but also broader design philosophies in LLM research: whether to retrofit explanations after the fact or build transparency directly into the model pipeline. Post-hoc tools often fail to capture the full complexity of internal decision pathways, especially under input perturbation or prompt rephrasing. This trade-off is central to the current challenges of XAI \cite{mumuni2025explainable}. This tension is particularly evident in high-risk domains, where trust and accountability are paramount. Their opacity limits their adoption in settings where explanation transparency is critical for oversight or compliance.

\subsection{Applicability: Toward Usable and Practical XAI}
\label{app:usable_xai}

Explainability for LLMs is often framed as a challenge making black-box systems more transparent and interpretable. However, a growing body of work argues that explainability should also be usable: practical roles in improving robustness, debugging workflows, safety monitoring, and user trust \cite{wu2403usable}. This shift reframes XAI not only as a mechanism for post-hoc transparency, but as a predictive, diagnostic, and adaptive tool throughout the LLM lifecycle.

Wu et al.\cite{wu2403usable} outline ten strategies for usable XAI, emphasizing explainability’s role across stages of model development, evaluation, and deployment. In this framework, LLMs are not merely targets of explainability, but also active agents in explanation processes. They can:

\begin{itemize}
    \item Generate explanations.
    \item Evaluate their own or other models’ outputs.
    \item Support interactive refinement with users of either the model outputs, the explanations themselves, or the evaluation criteria depending on the user role. For example, developers may iteratively refine evaluation criteria, while end users might interactively clarify or simplify the explanations provided by the model.
\end{itemize}

This perspective invites the design of explanation-aware architectures and workflows that treat XAI as integral, not peripheral, to LLM reliability and human-AI collaboration.

This shift also raises important accessibility considerations. For users with cognitive disabilities, or limited domain expertise, explanation usability hinges not just on clarity, but on adaptability. The ability of LLMs to interactively refine explanations presents a unique opportunity to support cognitive accessibility: they can simplify language, rephrase content, or structure reasoning incrementally, depending on user feedback. To fulfill this potential, explanation-aware systems must incorporate user modeling and interface accommodations that adapt explanation delivery to diverse cognitive needs. Thus, usable XAI must also be accessible to XAI.

\paragraph{Contrasting XAI needs from developers to end users.}
Explainability needs vary significantly depending on whether the consumer of the explanation is a developer tuning the system or an end-user relying on the system's output. Table~\ref{tab:developer_vs_user} summarizes these contrasting goals.

\begin{table}[ht]
\centering
\caption{Contrasting XAI needs for developers and end users}
\label{tab:developer_vs_user}
\begin{tabular}{|p{3cm}|p{5.5cm}|p{6cm}|}

\hline
\textbf{Dimension} & \textbf{Developer-Centric XAI} & \textbf{End User-Centric XAI} \\
\hline
Purpose & Debug, align, audit, monitor & Build trust, inform decisions, provide recourse \\
\hline
Explanation Style & Technical (tokens, gradients, circuits) & Narrative (rationales, analogies, conversational) \\
\hline
Tooling & Attribution maps, tracing tools, causal probes & Simplified rationales, counterfactuals, follow-up questions \\
\hline
Temporal Use & Continuous development and fine-tuning & On-demand, embedded in decision-making \\
\hline
Risk Sensitivity & Alignment, hallucinations, safety & Trustworthiness, fairness, justification \\
\hline
\end{tabular}
\end{table}

\paragraph{Pipeline Examples: Usable XAI in Practice.}
The usable XAI paradigm is exemplified in real-world LLM workflows. Below are three illustrative pipelines where explainability is tightly coupled to reliability and human trust:

\begin{enumerate}
    \item Hallucination Detection in Open-Domain Question-Answer:
In domain-specific Question-Answer (QA)  systems (e.g., biomedical), token-level attribution is used to detect hallucinations by highlighting unsupported answer tokens. Attribution heatmaps combined with retrieval confidence allow both models and users to identify when an output lacks grounding, improving factual precision.

\item  Safety Monitoring via XAI Filters:
XAI tools can monitor LLM safety by analyzing attention patterns or token gradients in response to prompts flagged as risky (e.g., jailbreak attempts). Models can self-explain refusal behavior or raise flags for human review when attention signals deviate from expected safe zones.

  \item  Explanation-Based Fine-Tuning Loops:
In feedback-driven development, LLMs generate self-rationales during supervised fine-tuning. These rationales are evaluated by humans (or simulated judges), and poor-quality explanations trigger either model updates or filtering of low-quality training data. This loop aligns the model reasoning with human expectations.
\end{enumerate}

\paragraph{Toward Explainability-Aware LLM Pipelines.}
Integrating explainability into production LLM pipelines as agentic LLMs requires modular interfaces, lightweight attribution methods, and adaptive feedback loops. Future work should focus on:

\begin{itemize}
    \item Toolkits that unify explanation generation, evaluation, and visualization.
    \item Metrics that link the XAI utility with downstream performance (for example, trust gain versus accuracy loss).
    \item Interfaces that present explanations in adjustable formats tailored to user roles and cognitive profiles.
\end{itemize}

In summary, usable XAI transforms explainability from an interpretive afterthought to a core enabler of reliability, alignment, and trust in LLM-powered systems.

\subsection{Opacity and Explainability at Scale}
\label{opacity}

Although post-hoc techniques, such as LIME, SHAP, or integrated gradients, remain dominant because of their model-agnostic flexibility, they often lack faithfulness when applied to large, non-linear architectures. Inherent interpretability instead embeds explanatory signals directly in the model’s reasoning, through approaches such as chain-of-thought prompting, concept bottleneck layers, or rationalization models that generate self-explanations \cite{mumuni2025explainable, cambria2024xai}.

Beyond tool design and developer workflows, explainability also raises normative questions: Who are the explanations meant for and how the residual opacity of large models should be governed.

This raises a deeper epistemic and normative question: is opacity always a flaw? Recent sociotechnical perspectives argue that opacity in foundation models should be treated not merely as a technical flaw, but as a \textit{governance condition}, one that must navigate through a stakeholder-calibrated explanation, institutional accountability, and pragmatic scaffolding \cite{herrera2025opacity}. Herrera and Calderón (2025) propose the \textit{Lack of Belief: Opacity \& eXplainability} (LoBOX) framework that considers opacity not as a flaw to be eliminated, but as a condition to be ethically governed. It advances a three-stage governance pathway: reducing accidental opacity, bounding irreducible opacity, and delegating trust through institutional oversight. LoBOX reframes explainability as a contextual, stakeholder-aligned obligation and positions institutional accountability as the foundation for public trust in AI systems, particularly in high-risk and inherently opaque environments. Khalili~\cite{khalili2024against} argues that opaque AI undermines users’ moral capacity and calls for a \textit{qualitative understanding} of AI, where people grasp its practical consequences rather than its computational innards. This perspective aligns with the \textbf{cognitive} and \textbf{causal} dimensions of TAXAL, while complementing LoBOX~\cite{herrera2025opacity}, which frames opacity as a governance condition. Together, they highlight the need for individual-level moral agency supported by institutional accountability.

Thus, in the era of foundation models, the objective of explainability is not total transparency, but responsible mediation. Rather than aiming for full transparency, which may be technically impossible, explainability should strive for \textit{justifiability}, \textit{contestability}, and \textit{trust calibration} within specific social roles and applications. For example, a physician using an LLM-powered diagnostic assistant needs a customized explanation that reflects clinical reasoning and liability standards, while a patient may require simplified justifications that support informed consent and trust in care.

This dual approach aligns with the definition of explainability by Arrieta et al.\cite{arrieta2020explainable} as a context-dependent and role-sensitive process. Herrera (2025) \cite{herrera2025reflections} emphasizes that effective XAI must evolve beyond visualizations or token saliency maps. Instead, it must facilitate \textit{cooperative reasoning} between humans and machines. This includes layered explanations, interfaces for user feedback, and mechanisms to resolve conflicts between model outputs and user values toward human-AI collaboration.

 This involves structuring accountability through mechanisms such as the LoBOX framework \cite{herrera2025opacity}, which combines legal scaffolds, procedural safeguards, and role-aware interfaces to ensure that models remain trustworthy and auditable even when fully interpretable mechanisms are beyond the reach.

\section{Explanation Strategies in Practice}
\label{app:strategies}

This appendix expands on subsection~\ref{sec:strategies}, providing a more detailed account of the current strategies used to explain and trust LLM behavior. Covers methods for generation, evaluation, and alignment. As we have discussed, each strategy can be mapped to one or more dimensions of the TAXAL framework, showing how the triadic fusion model connects conceptual design with practical implementation.

\paragraph{Natural Language Explanations and Self-Rationalization.}
LLMs can be prompted to generate natural language justifications for their predictions. This technique, known as self-rationalization, is in good agreement with user expectations and offers a low-friction form of explainability \cite{slack2023explaining}. However,  warn that these explanations may reflect post-hoc rationalization rather than genuine causal reasoning, reducing their faithfulness and reliability in critical domains.

\paragraph{Conversational and Interactive Interfaces.}
Systems such as \textit{TalkToModel} \cite{slack2023explaining} and the conversational agents studied by \cite{he2025conversational} represent a shift toward user-driven dialogic XAI. These interfaces allow users to ask follow-up questions, adjust the granularity of the explanation, and iteratively refine the output. They also support user feedback, helping to tailor the explanation to individual understanding levels and decision contexts.

\paragraph{Explainable Prompting and Training with LLM-Generated Rationale.}
Explanation-enriched prompting strategies, such as chain-of-thought (CoT) or knowledge-based templates, improve both performance and interpretability. \cite{wu2403usable} propose using explanations during training to scaffold model reasoning and to fine-tune smaller models with rationally enriched corpora. These strategies encourage models not only to predict accurately but also to explain their process transparently.

\paragraph{LLMs as Explanation Generators and Evaluators.}
LLMs can serve two roles in explainability pipelines, both as explanation generators and as automated plausibility evaluators. The evaluation mode is typically assessed through agreement with human-annotated rationales, benchmark datasets (e.g., e-SNLI, CoS-E), or meta-evaluation using consistency, coherence, and task-relevant scoring metrics. This enables novel workflows such as automated rationale annotation, plausibility scoring, or simulated domain expert feedback. Wu et al. \cite{wu2403usable} demonstrate that LLMs can emulate annotators in low-resource settings, assisting both in the training and evaluation of explanation models.

\paragraph{Leveraging Explainability for Model Improvement and Debugging.}
Explainability is not just diagnostic; it is increasingly used for targeted model refinement. Attribution and causal tracing methods have been applied to:

\begin{itemize}
    \item \textit{Edit factual knowledge} using methods such as ROME and MEMIT \cite{xu2024editing}.
    \item \textit{Enhance long-context reasoning} by analyzing attention drift and retrieval gaps.
    \item \textit{Reduce hallucinations} through attention-based interventions and saliency-guided filtering.
    \item \textit{Mitigate social bias} through interpretable gradient flows and counterfactual analysis.
\end{itemize}

These applications show that interpretability tools can be used not just to explain model decisions but also to fine-tune model behavior.

\paragraph{Explainability for Trust, Alignment, and Safety Prediction.}
Attribution-derived features (e.g., saliency scores or token importance vectors) can be used to train secondary classifiers that detect hallucination risk or safety violations. \cite{wu2403usable} present pipelines where attribution maps are inputted to lightweight models that predict when an LLM might produce harmful, false, or misaligned content. To ensure transparency and accountability, these lightweight models should themselves be explicable, adhere to the principles of XAI, and support inspection of their decision-making process. This transforms XAI into a predictive tool for monitoring trust and safety in real time.

\paragraph{Model Scale and the Limits of Post-Hoc Explanation.}
The size of the model plays a non-trivial role in the quality of the explanation. \cite{heyen2024effect} show that as models grow in scale, their internal logic becomes more traceable (increased faithfulness), but this does not translate into better alignment with human expectations (stable plausibility). This decoupling of performance from interpretability challenges the assumption that larger models are inherently more explainable.

\end{appendices}
\section*{Declaration of AI-assisted technologies in the writing process}

During the preparation of this work, the authors used large-language models to improve the readability and language of the manuscript. After using this tool/service, the authors reviewed and edited the content as needed and assumed full responsibility for the content of the published article.

\section*{Acknowledgements} This research results from the Strategic Project IAFER-Cib (C074/23), as a result of the collaboration agreement signed between the National Institute of Cybersecurity (INCIBE) and the University of Granada. This initiative is carried out within the framework of the Recovery, Transformation, and Resilience Plan funds, financed by the European Union (Next Generation).

\printbibliography

\end{document}